\documentclass[letterpaper, 10 pt, conference]{ieeeconf}  

\IEEEoverridecommandlockouts                              

\usepackage{epsfig} 
\usepackage{array}
\usepackage{booktabs}
\usepackage{makecell}
\usepackage{amsmath} 
\usepackage{amssymb}  
\usepackage{bm}
\usepackage[ruled,linesnumbered]{algorithm2e}
\usepackage{mathtools}
\usepackage[noend]{algpseudocode}
\usepackage{graphicx}
\usepackage{subfigure}
\usepackage{mwe}

\title{\LARGE \bf
Segmented Trajectory Optimization for Autonomous Parking in Unstructured Environments 
}

\author{Hang Yu and Renjie Li%
\thanks{All authors are with QCraft Inc., China, and contributed equally to this work.  
Emails: {\tt\small \{yuhang, renjie\}@qcraft.ai}.  
Corresponding author: Renjie Li ({\tt\small renjie@qcraft.ai}).}%
}

\begin{document}

\maketitle
\thispagestyle{empty}
\pagestyle{empty}

\begin{abstract}

This paper presents a Segmented Trajectory Optimization (STO) method for autonomous parking, which refines an initial trajectory into a dynamically feasible and collision-free one using an iterative SQP-based approach. STO maintains the maneuver strategy of the high-level global planner while allowing curvature discontinuities at switching points to improve maneuver efficiency. To ensure safety, a convex corridor is constructed via GJK-accelerated ellipse shrinking and expansion, serving as safety constraints in each iteration. Numerical simulations in perpendicular and reverse-angled parking scenarios demonstrate that STO enhances maneuver efficiency while ensuring safety. Moreover, computational performance confirms its practicality for real-world applications.

\end{abstract}

\section{INTRODUCTION} \label{sec.intro}
Autonomous driving is rapidly advancing, leading to its widespread adoption across various applications. As a key functionality, autonomous parking assists users in maneuvering a vehicle from an initial position into a selected parking spot. Although autonomous parking operates at lower speeds, it poses distinct challenges compared to on-road autonomous driving. The environment is often unstructured, cluttered, and constrained, with irregularly positioned obstacles \cite{Guo2023survey}.

Motion planning for autonomous parking is challenging. The planned trajectory must be dynamically feasible to ensure precise execution by the motion controller. It must also be strictly collision-free, accounting for the full geometry of the vehicle and obstacles. Additionally, parking maneuvers often involve multiple phases transitioned by gear shifts, and the planning algorithm must be computationally efficient to enable real-time implementation on embedded platforms.

Early motion planning studies for autonomous vehicles include sampling-based methods, such as RRT* \cite{Karaman2011sampling}, and search-based methods, such as hybrid A* \cite{Dolgov2010path}. While these methods can find near-optimal paths globally, the results are often non-smooth and fail to fully accommodate nonholonomic vehicle kinematics due to simplified vehicle models employed to mitigate the ``curse of dimensionality". For autonomous parking, early research primarily explored rule-based geometric methods for path generation \cite{Vorobieva2014automatic,Du2014autonomous}. As a simplified form of sampling-based planning, these approaches were designed for predefined standard parking scenarios, limiting their flexibility and applicability to general parking tasks.

Optimization-based motion planning formulates the problem as a constrained optimal control problem, incorporating system dynamics, motion smoothness, and safety constraints into a unified framework. Early applications include mobile robots \cite{Howard2010receding}, unmanned aerial vehicles (UAVs) \cite{Augugliaro2012generation}, robotic arms \cite{Schulman2013finding,Schulman2014motion}, and on-road autonomous driving \cite{Ziegler2014trajectory,Berntorp2014models}. In autonomous parking, optimization-based methods were first introduced in \cite{Li2015unified,Li2016time}. However, these methods face challenges due to nonlinear system dynamics and nonconvex collision-avoidance constraints. The quality of the optimized trajectory heavily depends on the initial guess, which determines the local feasible space. Zhang \emph{et al.} \cite{Zhang2018autonomous} proposed a hierarchical planning method, where Hybrid A*-generated coarse trajectory serves as the initial guess for trajectory optimization. However, the safety constraints remain nonconvex. To address this, several studies construct convex corridors around the initial trajectory as safety constraints\cite{Sun2021successive,Li2021optimization,Han2023efficient}. Despite these efforts, trajectory optimization for autonomous parking still faces three key challenges:

(1) \emph{Maneuver Efficiency}: Human drivers tend to steer aggressively at the beginning, during gear shifts, and when finalizing parking, resulting in curvature discontinuities at switching points (initial, terminal, and gear-shifting points). Search-based methods naturally allow such discontinuities, offering better maneuver efficiency. However, most optimization-based approaches enforce global curvature continuity \cite{Zhang2018autonomous,Sun2021successive,Li2021optimization}, potentially conflicting with the initial maneuver strategy. Han \emph{et al.} \cite{Han2023efficient} introduced a piece-wise polynomial trajectory optimization method that preserves the initial maneuver strategy but still maintains curvature continuity at switching points.

(2) \emph{Convex Corridor Construction}: Efficient and non-conservative convex corridor generation is crucial for trajectory optimization. Zhu \emph{et al.} \cite{Zhu2015convex} introduced the Convex Elastic Smoothing (CES) algorithm using circles, but it does not guarantee exact collision avoidance and is overly conservative. Liu \emph{et al.} \cite{Liu2017convex,Liu2018convex} proposed the Convex Feasible Set (CFS) approach using polygons, but constraint complexity scales with number of obstacles, potentially slowing convergence. Li \emph{et al.} \cite{Li2021optimization} employed axis-aligned boxes for faster optimization but at the cost of conservative space utilization. Deits \emph{et al.} \cite{Deits2015computing} introduced the Iterative Regional Inflation by Semidefinite programming (IRIS) method, which decomposes free space into convex polytopes via ellipsoidal construction, fully exploring the space but at high computational cost. Liu \emph{et al.} \cite{Liu2017planning} later improved IRIS with the Safe Flight Corridor (SFC) algorithm by accelerating ellipsoid computation, but its application is limited to occupancy obstacles and UAVs.

(3) \emph{Fast Optimization}: The nonlinear system dynamics and nonconvex collision-avoidance constraints pose significant computational challenges. To address this, sequential convex optimization techniques such as Sequential Quadratic Programming (SQP), have been widely applied in robotic motion planning \cite{Augugliaro2012generation,Schulman2013finding,Schulman2014motion}. These methods iteratively approximate the original problem by linearizing nonlinear constraints along the current trajectory. Each iteration solves a convex subproblem, progressively refining the solution until convergence. Several studies have adopted this iterative framework for autonomous trajectory optimization \cite{Sun2021successive,Li2021optimization,Liu2017convex,Liu2018convex,Bergman2018combining,Shi2019bilevel,Zhou2020autonomous}.

\emph{Contributions}: This paper presents a Segmented Trajectory Optimization (STO) method for autonomous parking, which refines a coarse initial trajectory into a dynamically feasible and collision-free trajectory. The main contributions include:

(1) \emph{Maneuver Efficiency}: STO strictly adheres to the maneuver strategy determined by the high-level global planner, ensuring that the optimized trajectory maintains the same maneuver phases as the initial trajectory. Additionally, it allows curvature discontinuities at switching points to enhance maneuver efficiency.

(2) \emph{Convex Corridor Construction}: Inspired by IRIS \cite{Deits2015computing} and SFC \cite{Liu2017planning}, convex polygon corridors are constructed through ellipse expansion to maximize space utilization. Unlike SFC, our approach leverages the Gilbert–Johnson–Keerthi (GJK) algorithm \cite{Gilbert2002fast} to efficiently compute the nearest point from general convex obstacles that have a uniform buffer to an ellipse.

(3) \emph{Fast Optimization}: Following the sequential convex optimization framework, we propose an iterative SQP-based algorithm to solve the STO problem efficiently. In each iteration, the trajectory is refined by solving an approximated convex subproblem derived from constraint linearization and convex corridor reconstruction.

The remainder of this paper is organized as follows: Section~\ref{sec.problem_formulation} formulates the STO problem. Section~\ref{sec.algorithm} details the proposed SQP-based algorithm. Section~\ref{sec.numerical_results} presents numerical simulations validating STO, and Section~\ref{sec.conclusions} concludes this paper.

\section{PROBLEM FORMULATION} \label{sec.problem_formulation}

This section formally defines the STO problem based on a reference trajectory and a set of buffered convex obstacles.

\subsection{Vehicle Geometry and Dynamics} \label{subsec.vehicle_geometry_and_dynamics}

\begin{figure}[thpb]
\centering
\includegraphics[scale = 1.3]{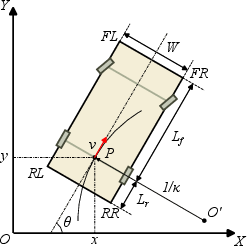}
\caption{Vehicle geometry and state. $P$: rear-axle center; $O'$: instantaneous center; $FL$: front-left corner; $FR$: front-right corner; $RL$: rear-left corner; $RR$: rear-right corner; $L_f$: distance from rear axle to front bumper; $L_r$: distance from rear axle to rear bumper; $W$: vehicle width.}
\label{fig.vehicle_geometry}
\end{figure}

The vehicle geometry and state representation are illustrated in Fig.~\ref{fig.vehicle_geometry}, where the vehicle is modeled as a rectangle. Following the kinematic bicycle model suitable for low-speed scenarios, the vehicle state is defined as $\mathbf{x}:=[x, y, \theta, v, \kappa]^\text{T}$ at the rear-axle center. Here, $\mathbf{p} := [x, y]^\text{T}$ represents the position, $\theta$ is the heading angle, $v$ is the longitudinal velocity, and $\kappa$ is the trajectory curvature. The control input is $\mathbf{u} := [a, \psi]^\text{T}$, where $a$ and $\psi$ represent the longitudinal acceleration and curvature rate, respectively. The system dynamics are given by:
\begin{equation} \label{eq.dynamics}
	\dot x      = v \cos\theta, \;\;
	\dot y      = v \sin\theta, \;\;
	\dot \theta = v \kappa, \;\;
	\dot v      = a, \;\;
	\dot \kappa = \psi
\end{equation}

Based on the vehicle geometry, the coordinates of the four corners can be computed as:
\begin{align} \label{eq.corners}
	\mathbf{p}_{FL,FR} & = \mathbf{p} + [L_f\cos\theta \mp \dfrac{1}{2} W \sin\theta, \,
	L_f\sin\theta \pm \dfrac{1}{2} W  \cos\theta]^\text{T} \notag                        \\
	\mathbf{p}_{RL,RR} & = \mathbf{p} - [L_r\cos\theta \pm \dfrac{1}{2} W \sin\theta, \,
	L_r\sin\theta \mp \dfrac{1}{2} W  \cos\theta]^\text{T}
\end{align}

\subsection{Obstacle Modeling} \label{subsec.obstacle_modeling}

\begin{figure}[thpb]
	\centering
	\includegraphics[scale = 1.3]{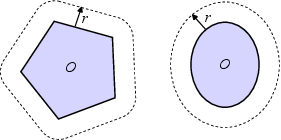}
	\caption{Convex obstacles with a uniform buffer.}
	\label{fig.buffered_obstacles}
\end{figure}

A uniform buffer is applied to obstacles to account for perception and control uncertainties. A convex obstacle $O$ with buffer $r$ forms the buffered obstacle $O_b$, which is defined as:
\begin{equation} \label{eq.obstacle_with_buffer}
	O_b := \left\{ \mathbf{o} + \mathbf{b} \;\middle|\; \mathbf{o} \in O, \; \mathbf{b} \in B \right\}
\end{equation}
where $B: = \left\{ \mathbf{p} \;\middle|\; \|\mathbf{p}\|_2 \leq r\right\}$ represents a disk centered at the origin with radius $r$. $O_b$ corresponds to the Minkowski sum of $O$ and $B$, or equivalently, can be visualized as the expanded region swept by a circle of radius $r$ along the perimeter of $O$. Fig.~\ref{fig.buffered_obstacles} provides two examples of buffered convex obstacles.

\subsection{Segmented Reference Trajectory} \label{subsec.segmented_reference_trajectory}

Consider a collision-free reference trajectory consisting of $M$ segments, each representing a forward or backward maneuver. Let the $i$-th segment contain $N_i$ points. The segmented reference trajectory is expressed as $\big\{ [\mathbf{x}_{r,0}^{(i)}, \dots, \mathbf{x}_{r,N_{i}-1}^{(i)}] \big\}_{i = 1}^{M}$. Each segment has a uniform velocity direction. Note that the initial reference trajectory need not be dynamically feasible, and the process of generating and refining it iteratively will be detailed in Section~\ref{subsec.overall_framework}.

\subsection{Constraints} \label{subsec.constraints}

Unless otherwise stated, the following constraints for the STO problem apply to $k = 0,\dots,N_i-1$ for the states, $k = 0,\dots,N_i-2$ for the control inputs, and to all $i = 1,\dots,M$.

\subsubsection{Kinematics}

The system dynamics in Eq.~\eqref{eq.dynamics} are discretized using the forward Euler method with timestep $T$, yielding the kinematic constraints:
\begin{equation} \label{eq.kinematics_constraint}
	\begin{bmatrix}
		x_{k+1}^{(i)}      \\
		y_{k+1}^{(i)}      \\
		\theta_{k+1}^{(i)} \\
		v_{k+1}^{(i)}      \\
		\kappa_{k+1}^{(i)}
	\end{bmatrix}
	=
	\begin{bmatrix}
		x_{k}^{(i)} + v_{k}^{(i)}\cos\theta_{k}^{(i)}\cdot T  \\
		y_{k}^{(i)} + v_{k}^{(i)}\sin\theta_{k}^{(i)}\cdot T  \\
		\theta_{k}^{(i)} + v_{k}^{(i)}\kappa_{k}^{(i)}\cdot T \\
		v_{k}^{(i)} + a_{k}^{(i)}\cdot T                      \\
		\kappa_{k}^{(i)} + \psi_{k}^{(i)}\cdot T
	\end{bmatrix}
\end{equation}
where $k$ ranges from 0 to $N_i-2$.

\subsubsection{Physical Limitation}

To ensure physical feasibility and comfort, the curvature, acceleration, and curvature rate are bounded as follows:
\begin{equation} \label{eq.physical_limit}
	\lvert \kappa_{k}^{(i)} \rvert \leq \kappa_{\text{max}}, \quad
	\lvert a_{k}^{(i)} \rvert \leq a_{\text{max}}, \quad
	\lvert \psi_{k}^{(i)} \rvert \leq \psi_{\text{max}}
\end{equation}
where $\kappa_{\text{max}}$, $a_{\text{max}}$ and $\psi_{\text{max}}$ are the corresponding bounds.

\subsubsection{Switching Points}

Curvature discontinuities are allowed at switching points, where curvature is not constrained but treated as a free optimization variable. Additionally, we assume each maneuver starts and ends at rest. The initial and terminal state constraints are given by:
\begin{equation} \label{eq.end_constraint}
\begin{aligned}
	 & x_{0}^{(1)}    = x_{s}, &  & y_{0}^{(1)} = y_{s},   &  & \theta_{0}^{(1)}  =\theta_{s},   &  & v_{0}^{(1)}    =0 \\
	 & x_{N_M-1}^{(M)}  = x_{f}, &  & y_{N_M-1}^{(M)} = y_{f}, &  & \theta_{N_M-1}^{(M)}  =\theta_{f}, &  & v_{N_M-1}^{(M)}  =0
\end{aligned}
\end{equation}
where $s$ and $f$ denotes the initial and terminal states, respectively. For gear-shifting points, the position, heading, and velocity constraints are:
\begin{equation}
\begin{aligned} \label{eq.gear_shift_constraint}
	 & x_{0}^{(i+1)} = x_{N_{i}-1}^{(i)},           &  & y_{0}^{(i+1)} = y_{N_{i}-1}^{(i)},    \\
	 & \theta_{0}^{(i+1)} = \theta_{N_{i}-1}^{(i)}, &  & v_{1}^{(i+1)} = v_{N_{i}-1}^{(i)} = 0
\end{aligned}
\end{equation}
where $i$ ranges from 1 to $M-1$.

\subsubsection{Reference Trajectory Proximity}

To minimize approximation errors in SQP iterations, the trajectory is required to stay close to the reference trajectory:
\begin{equation} \label{eq.proximity_constraint}
	\lvert \mathbf{p}_{k}^{(i)} - \mathbf{p}_{r, k}^{(i)} \rvert \leq \Delta \mathbf{p}_{\text{max}}, \quad
	\lvert \theta_{k}^{(i)} - \theta_{r, k}^{(i)} \rvert \leq \Delta \theta_{\text{max}}
\end{equation}
where $\Delta \mathbf{p}_{\text{max}}$ and $\Delta \theta_{\text{max}}$ are the proximity bounds for position and heading, respectively. Additionally, the velocity must be consistent with the reference trajectory direction and bounded:
\begin{equation} \label{eq.direction_constraint}
	\begin{cases}
		0 \leq v_{k}^{(i)} \leq v_{\text{max}}, & \text{if $i$-th segment is forward}  \\
		v_{\text{min}}\leq v_{k}^{(i)} \leq 0,  & \text{if $i$-th segment is backward}
	\end{cases}
\end{equation}
where $v_{\text{max}}$ and $v_{\text{min}}$ are the forward and backward velocity bounds, respectively.

\subsubsection{Collision Avoidance}

A safe convex corridor is constructed around the reference trajectory using convex polygons. Let $P_{k}^{(i)}$ be the polygon enclosing $\mathbf{x}_{r,k}^{(i)}$, expressed as $P_{k}^{(i)} = \big\{ \mathbf{p} \mid A_{k}^{(i)}\mathbf{p} \leq b_{k}^{(i)} \big\}$, where $A_{k}^{(i)} \in \mathbb{R}^{n \times 2}$ and $b_{k}^{(i)} \in \mathbb{R}^{n}$ are known matrices, and ${n}$ is the number of enclosing hyperplanes at this point. The method to construct the safe convex corridor will be presented in Section~\ref{subsec.convex_corridor_construction}. Collision avoidance is enforced by constraining all vehicle corners to remain within the polygon:
\begin{equation} \label{eq.safety_constraint}
	A_{k}^{(i)} \mathbf{p}_{\bullet, k}^{(i)} \leq b_{k}^{(i)} + \sigma_{k}^{(i)}\cdot \mathbf{1}, \sigma_{k}^{(i)} \geq 0, \text{for} \; \bullet \in \big\{ \text{FL, FR, RL, RR} \big\}
\end{equation}
where $\mathbf{p}_{\bullet, k}^{(i)}$ is computed from Eq.~\eqref{eq.corners}, and $\sigma_{k}^{(i)}$ is a slack variable ensuring solution feasibility.

\subsection{Objective Function} \label{subsec.objective_function}

The objective function for the STO problem is defined as:
\begin{equation} \label{eq.objective}
	\begin{aligned}
		\sum_{i = 1}^M \bigg[ \sum_{k=0}^{N_i-1} \Big( & w_{1}\big(x_{k}^{(i)}-x_{r,k}^{(i)}\big)^{2} + w_{2}\big(y_{k}^{(i)}-y_{r,k}^{(i)}\big)^{2}           \\
		                                       + & w_{3}\big(\theta_{k}^{(i)} -\theta_{r,k}^{(i)}\big)^{2} + w_{4}v_{k}^{(i)2} + w_{5}\kappa_{k}^{(i)2} \\
		                                      + & w_{6}\sigma_{k}^{(i)2} \Big)  + \sum_{k=0}^{N_i-2} \Big( w_{7}a_{k}^{(i)2} + w_{8}\psi_{k}^{(i)2}\Big)\bigg]
	\end{aligned}
\end{equation}
where $w_{1}$ to $w_{8}$ are cost weights. The first three terms ($w_{1}$ to $w_{3}$) penalize deviations from the reference trajectory, while the next two terms ($w_{4}$ and $w_{5}$) promote trajectory smoothness. The sixth term ($w_{6}$) penalizes violations of safety constraints, whereas the last two terms ($w_{7}$ and $w_{8}$) discourage excessive control inputs.

Thus, the STO problem is formally formulated as the optimization of the entire segmented trajectory $\bm{X} := \big\{ [\mathbf{x}_{0}^{(i)}, \dots, \mathbf{x}_{N_{i}-1}^{(i)}] \big\} _{i = 1}^{M}$, along with the corresponding control sequence, $\bm{U} := \big\{ [\mathbf{u}_{0}^{(i)}, \dots, \mathbf{u}_{{N_{i}-2}}^{(i)}] \big\} _{i = 1}^{M}$, subject to the constraints from \eqref{eq.kinematics_constraint} to \eqref{eq.safety_constraint}, and the objective function defined in \eqref{eq.objective}.

\section{ALGORITHM} \label{sec.algorithm}

This section describes the algorithm for solving the STO problem. Section~\ref{subsec.overall_framework} introduces the overall framework, while Section~\ref{subsec.convex_corridor_construction} details the method for constructing the safe convex corridor.

\subsection{Overall Framework} \label{subsec.overall_framework}

The STO problem is solved using an iterative SQP-based algorithm, as outlined in Algorithm~\ref{alg.sto}. The inputs include a set of buffered obstacles $\bm{O}_b$ and an initial collision-free path $\rho$, which extends from the initial state to the target state. This path can be generated by a high-level global path planner. In this study, we use Hybrid A* \cite{Dolgov2010path} to generate the initial path.

To initialize the process, we perform a simple trapezoidal speed planning on the path $\rho$ to generate the reference trajectory. The speed profile consists of three phases: constant acceleration, constant velocity, and constant deceleration.

In each iteration, we construct a convex corridor around the reference trajectory as the safety constraints. The nonlinear constraints \eqref{eq.kinematics_constraint} and \eqref{eq.safety_constraint} are then linearized via a first-order Taylor expansion along the reference trajectory, forming an approximate QP subproblem. After solving the QP, we evaluate the solution's dynamical feasibility using the 4th-order Runge-Kutta (RK4) method. If the feasibility error $e := \lvert \text{RK4}(\mathbf{x}_k, \mathbf{u}_k) - \mathbf{x}_{k+1} \rvert$ remains below a predefined tolerance $e_f$ along the trajectory, the iteration terminates, and the solution is returned as the optimized trajectory. Otherwise, the intermediate solution is used as the reference trajectory for the next iteration.

\begin{algorithm} \label{alg.sto}
	\caption{Segmented Trajectory Optimization}
	\KwIn{Buffered obstacles $\bm{O}_b$ and an initial path $\rho$}
	\KwOut{A feasible collision-free trajectory $\bm{X}$}
	Initialize vehicle geometry and algorithm parameters \\
	Set dynamical feasibility tolerance $e_{f}$ \\
	$\bm{X}_r \leftarrow$ PlanSimpleSpeed$(\rho)$ \\
	\While{true}{
		$\bm{P} \leftarrow$ ConstructConvexCorridor$(\bm{O}_b, \bm{X}_r)$ \\
		$\bm{X},\bm{U} \leftarrow$ SolveApproximateQP$(\bm{P}, \bm{X}_r)$ \\
		$ e \leftarrow$ EvalErrorWithRK4$(\bm{X},\bm{U})$ \\
		\If{ $ e < e_{f}$}{
			\textbf{break}
		}
		$\bm{X}_r \leftarrow \bm{X}$ 
	}
	\textbf{return} $\bm{X}$
\end{algorithm}

\subsection{Convex Corridor Construction} \label{subsec.convex_corridor_construction}

\subsubsection{Overall Process}

The convex corridor is defined as a series of connected convex polygons that enclose the reference trajectory while avoiding obstacles. Based on \cite{Deits2015computing} and \cite{Liu2017planning}, the convex polygon for each reference trajectory point is constructed in two steps: ellipse generation and polygon generation.

\textbf{\textit{Ellipse Generation (Step 1)}}: An initial ellipse is created at the reference trajectory point, with the center at the vehicle's center and the major axis aligned with the vehicle's heading. The major-to-minor axis ratio is set according to the vehicle's length-to-width ratio.

First, the length of the major axis is adjusted to ensure it does not collide with obstacles. Then, for each obstacle, the closest point to the ellipse is determined (\emph{i.e.}, the point with the minimum distance to the ellipse center in the uniform transformation space). If this point lies inside the ellipse, the minor axis is shortened to make the ellipse just touch the closest point. This process is repeated for all obstacles, resulting in the final ellipse.

\textbf{\textit{Polygon Generation (Step 2)}}: Starting with the ellipse obtained in the previous step, the closest point among all obstacles is identified. The ellipse is uniformly expanded until it just touches this point. A hyperplane is then constructed at the tangent to the expanded ellipse, defining a supporting half-space that contains the ellipse. Obstacles outside this half-space are removed or clipped at the boundary. This process is repeated until all obstacles are eliminated. The intersection of all constructed half-spaces forms the final convex polygon.

\begin{figure*} [thpb]
    \centering
    \subfigure[The obstacles are iterated in the order: pentagon, ellipse, and square. The initial red ellipse is shrunk to touch the closest point $\mathbf{p}_{1*}$ of the pentagon (blue ellipse). Then, it is further shrunk to touch $\mathbf{p}_{2*}$ of the ellipse (green ellipse). No shrinking occurs for the square, as its closest point lies outside the green ellipse. ]{
	\includegraphics[width=0.35\textwidth]{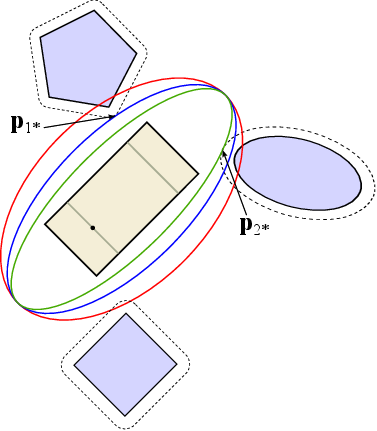}
        \label{fig.ellipse_shrinking}
    }
    \hspace{1.0cm}
    \subfigure[The initial red ellipse touches the closest point of the pentagon, forming a tangent hyperplane that eliminates obstacles outside the half-space. The ellipse expands to touch the closest points of the ellipse (blue) and the square (green), each time eliminating obstacles outside the new half-space. The intersection of all half-spaces forms the final polygon. ]{
	\includegraphics[width=0.4\textwidth]{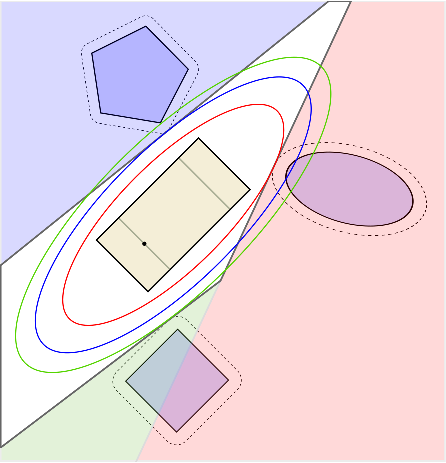}
        \label{fig.ellipse_expansion}
    }
    \caption{Corridor construction process. (a) Step 1: Ellipse generation; (b) Step 2: Polygon generation.}
    \label{fig.corridor_construction}
\end{figure*}

The convex corridor construction process for a single trajectory point is summarized in Algorithm~\ref{alg.corridor}, while Fig.~\ref{fig.corridor_construction} illustrates an example. 

\begin{algorithm} \label{alg.corridor}
	\caption{Convex Corridor Construction}
	\KwIn{A set of buffered obstacles $\bm{O}_b$ and a reference trajectory point $\mathbf{x}_r$}
	\KwOut{A safe convex polygon $P$ enclosing $\mathbf{x}_r$}
	\tcc{Step 1: Ellipse generation.}
	Initialize an ellipse $\varepsilon$ at $\mathbf{x}_r$ \\
	\For{$O_b \in \bm{O}_b$}{
		$\mathbf{p}_{*} \leftarrow$ FindClosestPoint$(O_b)$\\
		\If{$\mathbf{p}_{*}$ in $\varepsilon$}{
			$\varepsilon \leftarrow$ ShortenMinorAxis$(\varepsilon, \mathbf{p}_{*})$\\
		}
	}
	\tcc{Step 2: Polygon generation.}
	$\bm{O}_{b,\text{remain}} \leftarrow \bm{O}_b$\\
	$P \leftarrow \mathbb{R}^{2}$ \\
	\While {$\bm{O}_{b,\text{remain}} \neq \emptyset$}{
	$\mathbf{p}_{*} \leftarrow$ FindClosestPoint$(\bm{O}_{b,\text{remain}})$ \\
	$\varepsilon \leftarrow$ ExpandEllipse$(\varepsilon, \mathbf{p}_{*})$ \\
	$H \leftarrow$ ConstructHalfSpace$(\varepsilon, \mathbf{p}_{*})$ \\
	$\bm{O}_{b,\text{remain}} \leftarrow$ RemoveOrClipObstacles$(\bm{O}_{b,\text{remain}}, H)$ \\
	$P \leftarrow P \cap H$
	}
	\textbf{return} $P$
\end{algorithm}

\subsubsection{Calculation Details}

We present the calculation details in Algorithm~\ref{alg.corridor}, with a particular focus on accelerating the closest point computation using the GJK algorithm. A general ellipse is expressed as:
\begin{equation} \label{eq.ellipse_def}
	\varepsilon := \big\{ C\mathbf{p} + \mathbf{d} \mid \|\mathbf{p}\|_{2} \leq 1 \big\}
\end{equation}
where $\mathbf{d}$ represents the center of the ellipse. The matrix $C$ can be decomposed as $C = R\Lambda R^{\text{T}}$, where $R$ is a rotation matrix, and $\Lambda = \text{diag}(\alpha, \beta) $, with $\alpha$ and $\beta$ denoting the semi-major and semi-minor axes, respectively. We define an affine transformation as $\mathcal{F}(\mathbf{p}) := C\mathbf{p} + \mathbf{d}$. The ellipse can be uniformly scaled by the following transformation:
\begin{equation} \label{eq.ellipse}
	\varepsilon_{\gamma} := \big\{\mathcal{F}(\mathbf{p}) \mid \|\mathbf{p}\|_{2} \leq \gamma \big\}
\end{equation}
where $\gamma > 0$ is the scaling factor.

The problem of finding the closest point of a buffered obstacle to the ellipse in the uniform transformation space is formulated as:
\begin{equation} \label{eq.closest_point}
	\begin{aligned}
		\mathbf{p}_{*} = \underset{\mathbf{p}\in O_b, \mathbf{p}\in \varepsilon_{\gamma}}{\arg\min} & \; \gamma \\
	\end{aligned}
\end{equation}
This can be interpreted as determining the smallest $\varepsilon_{\gamma}$ such that $\mathbf{p}_{*} \in \varepsilon_{\gamma} \cap O_b \neq \emptyset$. Instead of solving this problem directly, we reformulate it in the inverse transformation space. The inverse transformation of $\mathcal{F}$ is defined as $\mathcal{F}^{-1}(\mathbf{p}) := C^{-1}(\mathbf{p} - \mathbf{d})$, which is also an affine transformation. Consequently, the inverse transformation of ellipse $\varepsilon_{\gamma}$ is expressed as:
\begin{equation} \label{eq.inverse_ellipse}
	\mathcal{F}^{-1}(\varepsilon_{\gamma}) = \big\{\hat{\mathbf{p}} \mid \|\hat{\mathbf{p}}\|_{2} \leq \gamma \big\}
\end{equation}
which represents a circle centered at the origin with radius $\gamma$. The problem in \eqref{eq.closest_point} is equivalent to finding the smallest circle $\mathcal{F}^{-1}(\varepsilon_{\gamma})$ such that $\hat{\mathbf{p}}_{*} \in \mathcal{F}^{-1}(\varepsilon_{\gamma}) \cap \mathcal{F}^{-1}(O_b) \neq \emptyset$. Given the definition of $O_b$ in \eqref{eq.obstacle_with_buffer}, $\mathcal{F}^{-1}(O_b)$ can be expressed as $\mathcal{F}^{-1}(O) + C^{-1}(B)$. Thus, the closest point finding problem in the inverse space can be formulated as:
\begin{equation} \label{eq.closest_point_inverse}
	\begin{aligned}
		\hat{\mathbf{p}}_{*} = \underset{\hat{\mathbf{p}}}{\arg\min} & \quad\; \|\hat{\mathbf{p}}\|_2 \\
		\text{s.t.} \quad & \; \hat{\mathbf{p}} = \hat{\mathbf{o}} + \hat{\mathbf{b}} \\
			    & \; \hat{\mathbf{o}} \in \mathcal{F}^{-1}(O) \\
			    & \; \hat{\mathbf{b}} \in C^{-1}(B)
	\end{aligned}
\end{equation}
Once $\hat{\mathbf{p}}_{*}$ is obtained, the original closest point can be recovered as $\mathbf{p}_{*} = \mathcal{F}(\hat{\mathbf{p}}_{*})$. The optimization problem \eqref{eq.closest_point_inverse} is convex, as both $\mathcal{F}^{-1}(O)$ and $C^{-1}(B)$ are convex sets, and it can be solved using general convex optimization techniques. However, solving it this way may be computationally expensive. To improve efficiency, we leverage the GJK algorithm, which iteratively determines the support (extreme) point in a given direction, updates the search direction, and refines the simplex to approximate the closest point, ensuring fast convergence. For simplicity, the support function to search the support point for any convex set $\bullet$ in a direction $\mathbf{v}$ is defined as:
\begin{equation}
	S(\bullet, \mathbf{v}) := \underset{\mathbf{p} \in \bullet}{\arg\max} \; \langle \mathbf{p}, \mathbf{v}\rangle
\end{equation}
For GJK, we assume that $S(\mathcal{F}^{-1}(O), \mathbf{v})$ and $S(C^{-1}(B), \mathbf{v})$ are implemented. Since the Minkowski sum property holds for support functions, it follows that $S(\mathcal{F}^{-1}(O_b), \mathbf{v}) = S(\mathcal{F}^{-1}(O), \mathbf{v}) + S(C^{-1}(B), \mathbf{v})$. With these, the GJK-based closest point finding algorithm can be carried out as in Algorithm~\ref{alg.gjk_algorithm}. For further details on the GJK algorithm, we refer the reader to \cite{Gilbert2002fast}.
\begin{algorithm}
	\caption{FindClosestPointWithGJK}\label{alg.gjk_algorithm}
	\KwIn{An inverse buffered obstacle $\mathcal{F}^{-1}(O_b)$}
	\KwOut{$\hat{\mathbf{p}}_{*} \in \mathcal{F}^{-1}(O_b)$ closest to origin}
	Set GJK tolerance $e_\text{max}$\\
	$\hat{\mathbf{p}}_{*} \leftarrow$ SelectArbitraryPointIn$(\mathcal{F}^{-1}(O_b))$\\
	Initialize simplex $\varsigma \leftarrow \{\hat{\mathbf{p}}_{*}\}$\\
	\While{true}{
		$\mathbf{w} \leftarrow$ $S(\mathcal{F}^{-1}(O_b), -\hat{\mathbf{p}}_{*})$ \\
		\If{$\lvert \|\hat{\mathbf{p}}_{*}\|_2 - \langle \hat{\mathbf{p}}_{*}, \mathbf{w} \rangle \rvert < e_\text{max}$}{
			\textbf{return} $\hat{\mathbf{p}}_{*}$
		}
		$\varsigma \leftarrow \{w\} \cup \varsigma$ \\
		$\hat{\mathbf{p}}_{*}, \varsigma \leftarrow $ FindClosestPoint$(\varsigma)$ \\
		$\varsigma \leftarrow$ UpdateSimplex$(\varsigma)$ \\
		\If{$\|p_{*}\|_2 < e_\text{max}$}{
                        $\hat{\mathbf{p}}_{*} \leftarrow$ origin \\
			\textbf{return} $\hat{\mathbf{p}}_{*}$
		}
	}
\end{algorithm}

In the ellipse generation step, once the closest point in inverse space $\hat{\mathbf{p}}_{*}$ is found, we can quickly verify whether the original point $\mathbf{p}_{*} = \mathcal{F}(\hat{\mathbf{p}}_{*})$ lies inside the original ellipse by checking $\|\hat{\mathbf{p}}_{*}\|_2 < 1$. The minor axis is then updated by substituting $\mathbf{p}_{*}$ into the ellipse equation while keeping the major axis fixed.

For hyperplane construction, the expansion of the ellipse to touch $\mathbf{p}_{*}$ is achieved by setting $\gamma = \|\hat{\mathbf{p}}_{*}\|_2$ in Eq.~\eqref{eq.ellipse}. The supporting tangent half-space at $\mathbf{p}_{*}$ is given by $H = \big\{ \mathbf{p} \mid \mathbf{a}^{T}\mathbf{p} \leq b \big\}$, where $\mathbf{a} = C^{-1}C^{-\text{T}}(\mathbf{p}_{*} - \mathbf{d})$ and $b = \mathbf{a}^{\text{T}}\mathbf{p}_{*}$.

\section{NUMERICAL RESULTS} \label{sec.numerical_results}

\subsection{Simulation Setup} \label{subsec.simulation_setup}

The simulations were performed on an Ubuntu 22.04 PC with an AMD Ryzen 9 5950X CPU (3.4 GHz). The proposed method was implemented in C++17, with the quadratic programming problem in each SQP iteration solved via the HPIPM solver \cite{Frison2020hpipm}. The key simulation parameters, including vehicle geometry and algorithm settings, are provided in Table~\ref{tab.simulation_parameters}.

To validate the effectiveness of our approach, we designed two parking simulation scenarios. The first scenario represents a standard perpendicular parking case (see Fig.~\ref{fig.vert}), where the target parking space is flanked by two parked vehicles, with a wall at the rear and another wall directly opposite, thereby restricting maneuverability. The second scenario represents a more challenging reverse-angled parking case (see Fig.~\ref{fig.reserve_sto}), which requires more maneuvers. An additional parked vehicle to the left of the target space further limits maneuverability.

For the perpendicular parking scenario, the initial state is defined as $x_{s} = 0$, $y_{s} = 0$, $\theta_{s} = 0$, with the target state set to $x_{f} = -3.7$, $y_{f} = -3.7$, $\theta_{f} = -1.6$. In the reverse-angled parking scenario, the initial state is set to $x_{s} = 0$, $y_{s} = 0$, $\theta_{s} = \pi$, and the final state to $x_{f} = -6$, $y_{f} = 0.6$, $\theta_{f} = \pi/4$. In both scenarios, all obstacle buffers are set to 0.2 m.

\begin{table}
\centering
\normalsize
\caption{Simulation Parameters}
\label{tab.simulation_parameters}
\resizebox{\columnwidth}{!}{
\begin{tabular}{cll}
\toprule
Parameter & \multicolumn{1}{c}{Description} & \multicolumn{1}{c}{Value} \\
\midrule
$L_f$ & rear axle to front bumper & 3.89 m \\
$L_r$ & rear axle to rear bumper & 1.043 m \\
$W$ & vehicle width & 1.87 m \\
$\kappa_{\text{max}}$ & maximum curvature & 0.16 m$^{-1}$ \\
$a_{\text{max}}$ & maximum acceleration & 1 ms$^{-2}$ \\
$\psi_{\text{max}}$ & maximum curvature rate & 0.03 m$^{-1}$s$^{-1}$ \\
$v_{\text{max}},\,v_{\text{min}}$ & velocity bounds & $\pm$3 ms$^{-1}$, \\
$\Delta \mathbf{p}_{\text{max}}$ & position proximity bound & [3, 3] m \\
$\Delta \theta_{\text{max}}$ & heading proximity bound & 0.175 rad \\
$w_{1}$, $w_{2}$, $w_{3}$, $w_{4}$ & cost weights & 0.3, 0.3, 0.1, 1.8 \\
$w_{5}$, $w_{6}$, $w_{7}$, $w_{8}$ & cost weights & 30, 10, 5, 100 \\
$e_\text{max}$ & GJK tolerance & 10$^{-8}$ \\
$e_f$ & feasibility tolerance & \makecell[l]{[0.01, 0.01, 0.01, \\ 10$^{-4}$,10$^{-4}$]} \\
\bottomrule
\end{tabular}
}
\end{table}

\subsection{Results and Analysis} \label{subsec.results}

\subsubsection{Feasibility and Safety}

\begin{figure}
    \centering
    \subfigure[Planned trajectory.]{
        \includegraphics[width=0.41\textwidth,trim={26 20 37 37},clip]{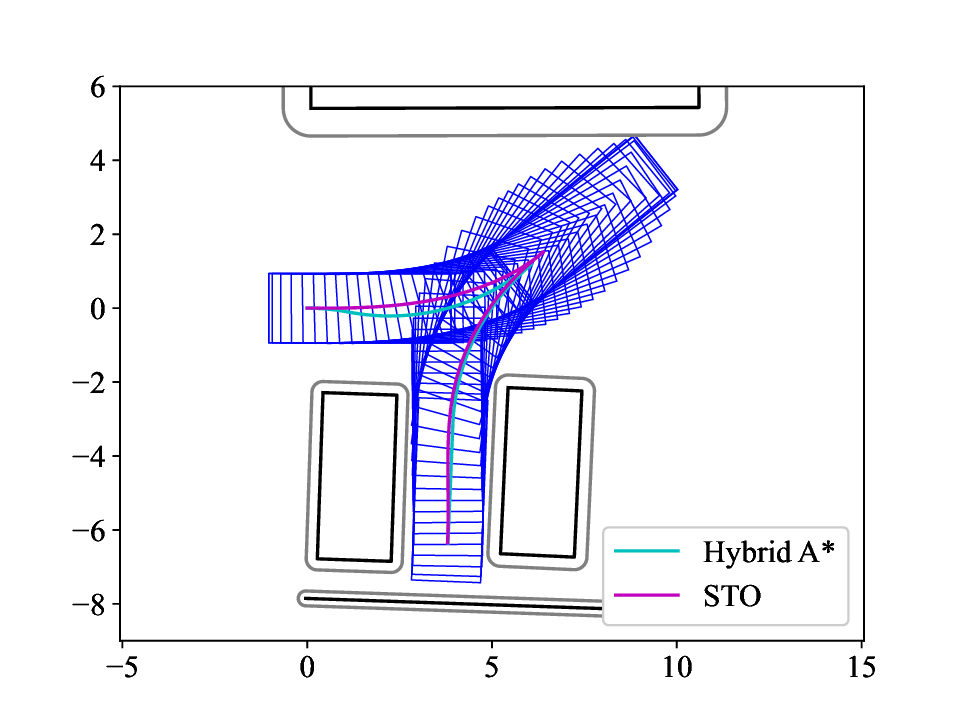}
        \label{fig.vert.contour}
    }
    \subfigure[Constructed convex corridor.]{
        \includegraphics[width=0.41\textwidth,trim={26 20 37 37},clip]{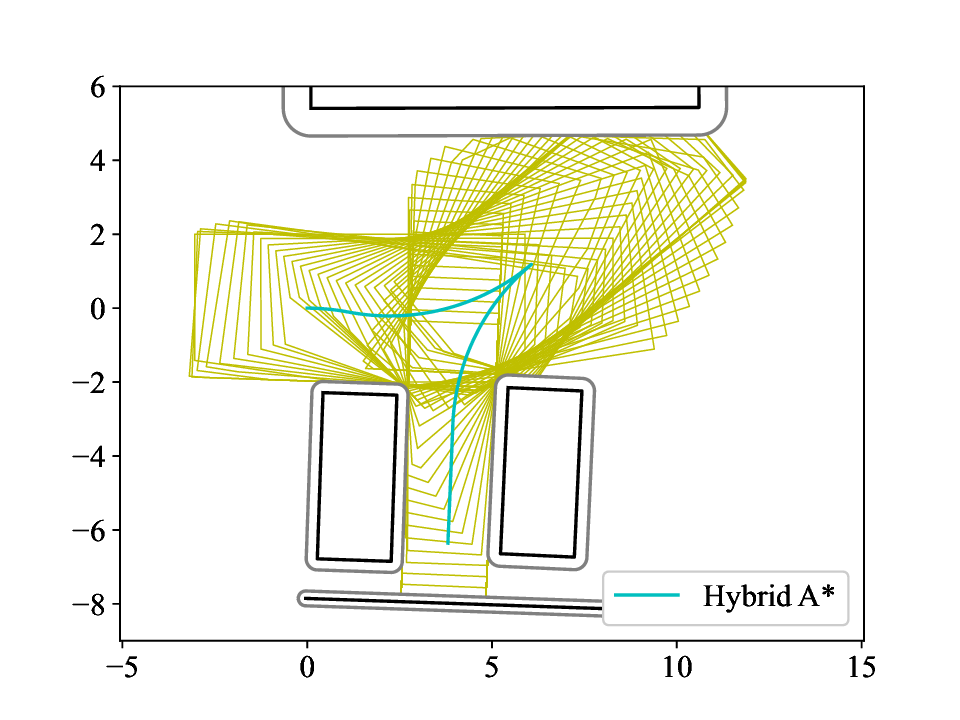}
        \label{fig.vert.convex_corridor}
    }
    \subfigure[Curvature and curvature rate.]{
        \includegraphics[width=0.41\textwidth,trim={26 20 37 37},clip]{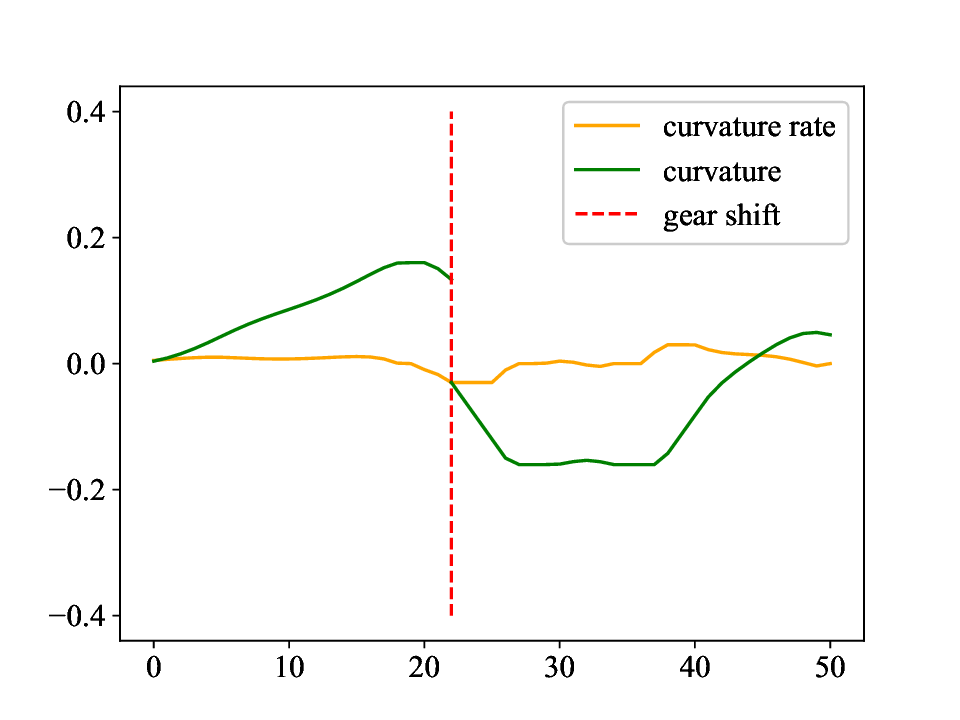}
        \label{fig.vert.kp}
    }
    \caption{Perpendicular parking.}
    \label{fig.vert}
\end{figure}

Fig.~\ref{fig.vert} presents the results for the perpendicular parking scenario. In Fig.~\ref{fig.vert}(a), the initial coarse trajectory generated by Hybrid A* is refined using our STO method while ensuring a safe distance from the buffered obstacles. Fig.~\ref{fig.vert}(b) illustrates the convex corridor constructed around the initial trajectory. It is clear from the figure that the corridor remains free of the buffered obstacles, thereby guaranteeing the safety of the refined trajectory. Fig.~\ref{fig.vert}(c) demonstrates the dynamical feasibility of the refined trajectory, showing that the curvature and curvature rate are within the predefined bounds and smooth for each maneuver. Notably, the curvature exhibits an abrupt change at the gear shift, indicating that our method enables rapid steering adjustments during phase transitions.

\subsubsection{Maneuver Efficiency}

\begin{figure} [t]
    \centering
    \subfigure[Planned trajectory.]{
        \includegraphics[width=0.41\textwidth,trim={26 20 37 37},clip]{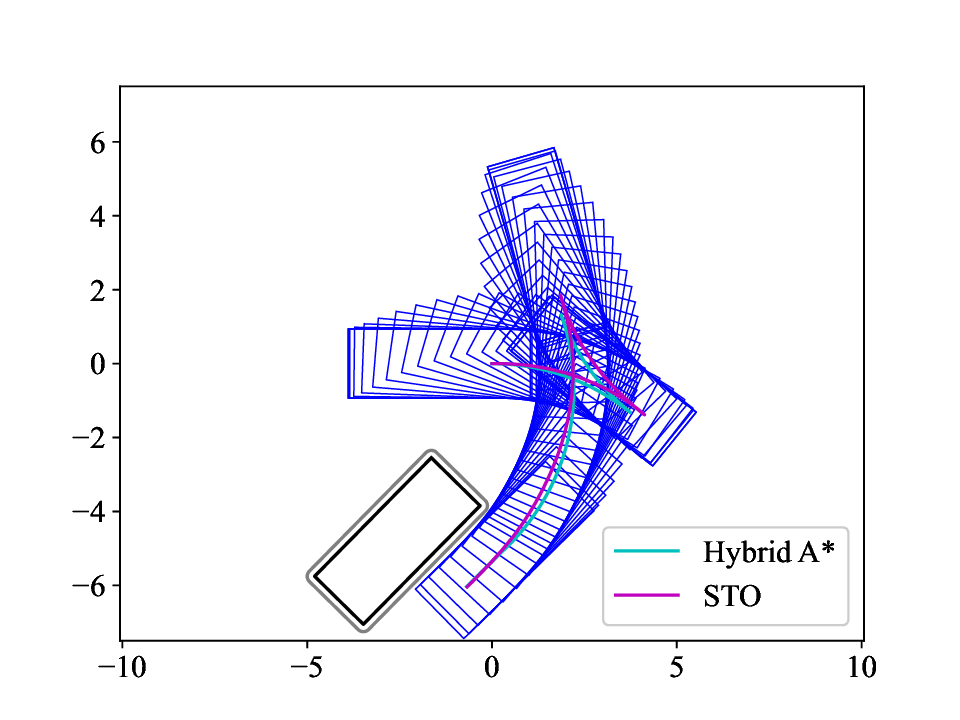}
        \label{fig.reverse.path}
    }
    \subfigure[Curvature and curvature rate.]{
        \includegraphics[width=0.41\textwidth,trim={26 20 37 37},clip]{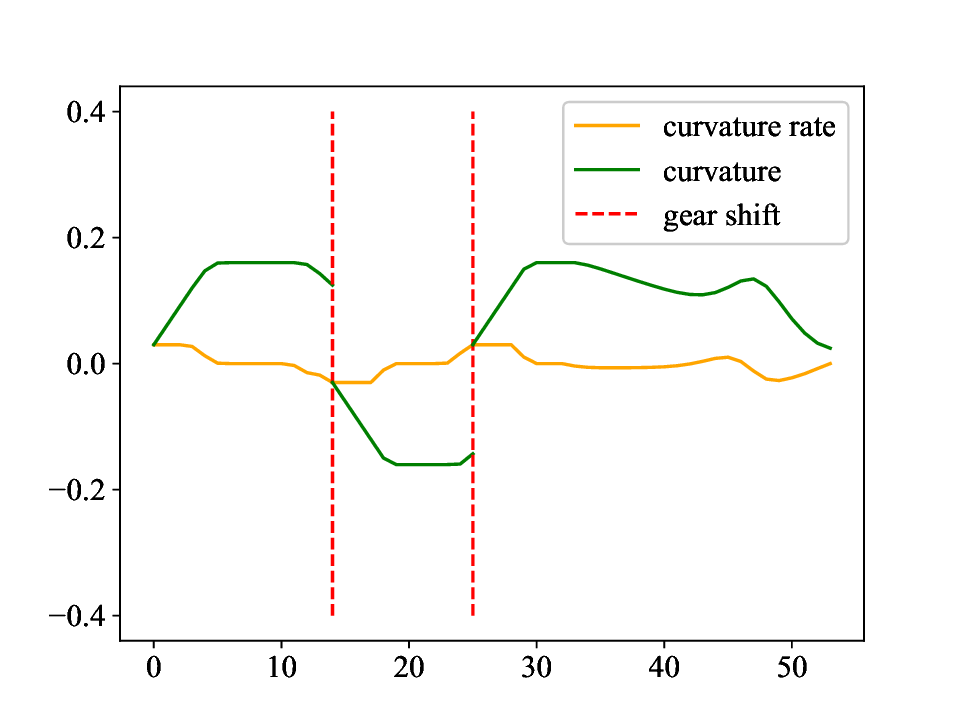}
        \label{fig.reverse.kp}
    }
    \caption{Reserve-angled parking: STO.}
    \label{fig.reserve_sto}
\end{figure}

\begin{figure} [t]
    \centering
    \subfigure[Planned trajectory.]{
        \includegraphics[width=0.41\textwidth,trim={26 20 37 37},clip]{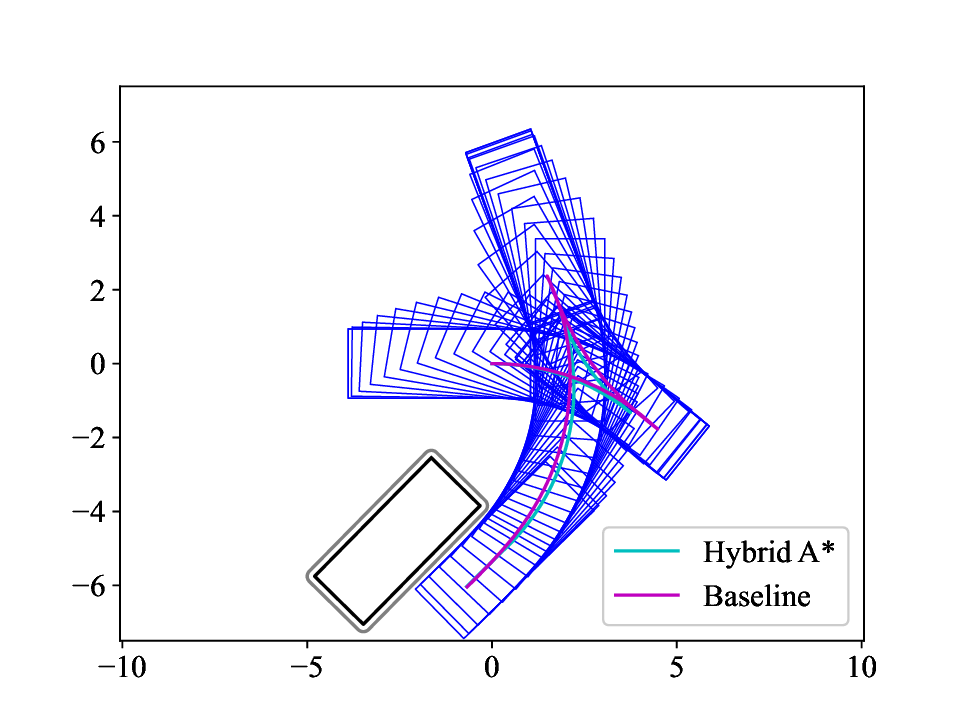}
        \label{fig.reverse.path_baseline}
    }
    \subfigure[Curvature and curvature rate.]{
        \includegraphics[width=0.41\textwidth,trim={26 20 37 37},clip]{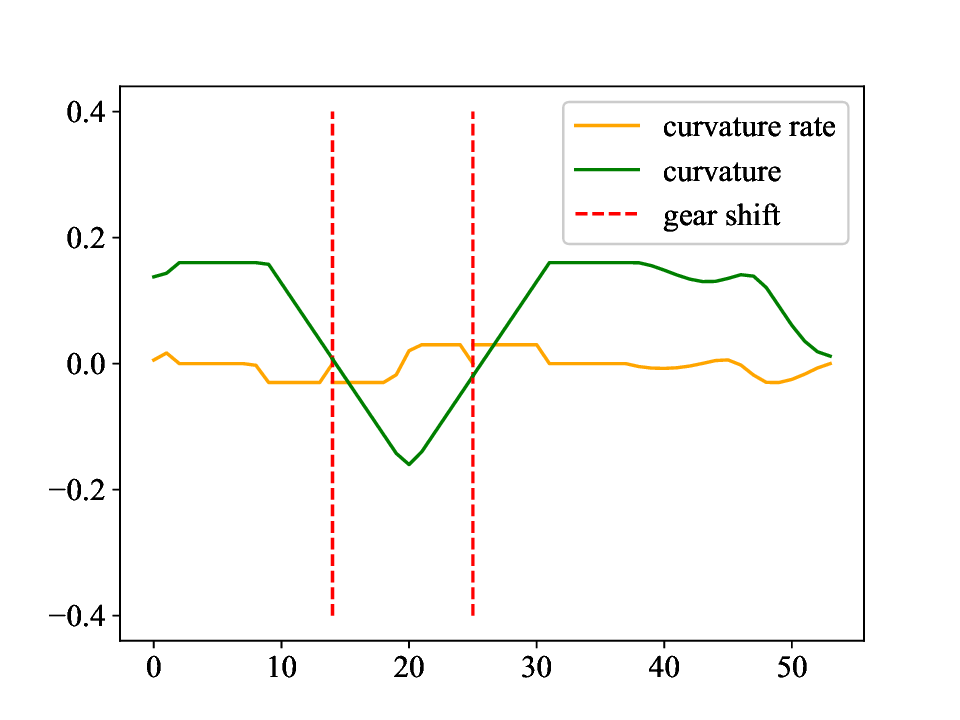}
        \label{fig.reverse.kp_baseline}
    }
    \caption{Reserve-angled parking: baseline.}
    \label{fig.reverse_baseline}
\end{figure}

To evaluate the maneuver efficiency of our approach, we compare STO with a baseline method in the reversed-angle parking scenario. The baseline method is a modified version of STO, where an additional curvature equality constraint is enforced at gear-shifting points, thereby preventing curvature discontinuities. Figures~\ref{fig.reserve_sto} and \ref{fig.reverse_baseline} illustrate the results of STO and the baseline method, respectively. By comparing Figure~\ref{fig.reserve_sto}(a) and \ref{fig.reverse_baseline}(a), it is evident that our method yields a shorter maneuvering distance. This confirms that allowing curvature discontinuities at gear shifts enhances maneuver efficiency. Quantitatively, the trajectory produced by STO (8.87 m) is 6\% shorter than that of the baseline method (9.43 m). Figures~\ref{fig.reserve_sto}(b) and \ref{fig.reverse_baseline}(b) further highlight the distinct curvature continuity properties of the two approaches at gear shifts. Fig.~\ref{fig.reserve_sto}(a) also shows that the maneuvering strategy provided by the global Hybrid A* planner is strictly maintained, as both trajectories follow the same maneuvering sequence.

\subsubsection{Computational Performance}

Table~\ref{tab.computational_performance} summarizes the computational performance of STO in the two parking scenarios. In both cases, STO converges within two iterations, with a total computation time under 10 ms. Notably, the convex corridor construction is especially fast, benefiting from GJK acceleration. These results demonstrate the efficiency and practical applicability of our method for real-world implementations.

\begin{table}
\centering
\normalsize
\caption{Computational Performance}
\label{tab.computational_performance}
\resizebox{\columnwidth}{!}{
\begin{tabular}{lcc}
\toprule
\multicolumn{1}{c}{Scenario} & Perp. Parking & Revers. Parking \\
\midrule
convergence iterations & 2 & 2 \\
corridor construction time & 0.910 ms & 0.454 ms \\
optimization time & 3.869 ms & 5.737 ms \\
total time & 4.779 ms & 6.191 ms \\
\bottomrule
\end{tabular}
}
\end{table}

\section{CONCLUSIONS} \label{sec.conclusions}

This paper presents a Segmented Trajectory Optimization (STO) method for refining an initial trajectory into a dynamically feasible and collision-free one for autonomous parking in unstructured environments. STO adheres to the maneuver strategy defined by the high-level global planner, while allowing for curvature discontinuities at the switching points to enhance maneuver efficiency. To address the computational challenges of STO, we propose an iterative SQP-based algorithm that enforces safety constraints through a convex corridor, which is constructed via GJK-accelerated ellipse shrinking and expansion.

Numerical simulations conducted in perpendicular and reverse-angled parking scenarios demonstrate that STO not only improves maneuver efficiency but also guarantees safety in maneuverability-limited environments. Additionally, computational performance analysis confirms that the proposed algorithm converges rapidly within a reasonable time, validating its practicality for real-world applications.

\addtolength{\textheight}{-12cm}   


\bibliographystyle{ieeetr}
\bibliography{ref}

\begin{thebibliography}{10}

\bibitem{Guo2023survey}
Y.~Guo, Z.~Guo, Y.~Wang, D.~Yao, B.~Li, and L.~Li, ``A survey of trajectory
  planning methods for autonomous driving—part i: Unstructured scenarios,''
  {\em IEEE Transactions on Intelligent Vehicles}, 2023.

\bibitem{Karaman2011sampling}
S.~Karaman and E.~Frazzoli, ``Sampling-based algorithms for optimal motion
  planning,'' {\em The international journal of robotics research}, vol.~30,
  no.~7, pp.~846--894, 2011.

\bibitem{Dolgov2010path}
D.~Dolgov, S.~Thrun, M.~Montemerlo, and J.~Diebel, ``Path planning for
  autonomous vehicles in unknown semi-structured environments,'' {\em The
  international journal of robotics research}, vol.~29, no.~5, pp.~485--501,
  2010.

\bibitem{Vorobieva2014automatic}
H.~Vorobieva, S.~Glaser, N.~Minoiu-Enache, and S.~Mammar, ``Automatic parallel
  parking in tiny spots: Path planning and control,'' {\em IEEE Transactions on
  Intelligent Transportation Systems}, vol.~16, no.~1, pp.~396--410, 2014.

\bibitem{Du2014autonomous}
X.~Du and K.~K. Tan, ``Autonomous reverse parking system based on robust path
  generation and improved sliding mode control,'' {\em IEEE Transactions on
  Intelligent Transportation Systems}, vol.~16, no.~3, pp.~1225--1237, 2014.

\bibitem{Howard2010receding}
T.~M. Howard, C.~J. Green, and A.~Kelly, ``Receding horizon model-predictive
  control for mobile robot navigation of intricate paths,'' in {\em Field and
  Service Robotics: Results of the 7th International Conference}, pp.~69--78,
  Springer, 2010.

\bibitem{Augugliaro2012generation}
F.~Augugliaro, A.~P. Schoellig, and R.~D'Andrea, ``Generation of collision-free
  trajectories for a quadrocopter fleet: A sequential convex programming
  approach,'' in {\em 2012 IEEE/RSJ international conference on Intelligent
  Robots and Systems}, pp.~1917--1922, IEEE, 2012.

\bibitem{Schulman2013finding}
J.~Schulman, J.~Ho, A.~X. Lee, I.~Awwal, H.~Bradlow, and P.~Abbeel, ``Finding
  locally optimal, collision-free trajectories with sequential convex
  optimization.,'' in {\em Robotics: science and systems}, vol.~9, pp.~1--10,
  Berlin, Germany, 2013.

\bibitem{Schulman2014motion}
J.~Schulman, Y.~Duan, J.~Ho, A.~Lee, I.~Awwal, H.~Bradlow, J.~Pan, S.~Patil,
  K.~Goldberg, and P.~Abbeel, ``Motion planning with sequential convex
  optimization and convex collision checking,'' {\em The International Journal
  of Robotics Research}, vol.~33, no.~9, pp.~1251--1270, 2014.

\bibitem{Ziegler2014trajectory}
J.~Ziegler, P.~Bender, T.~Dang, and C.~Stiller, ``Trajectory planning for
  bertha—a local, continuous method,'' in {\em 2014 IEEE intelligent vehicles
  symposium proceedings}, pp.~450--457, IEEE, 2014.

\bibitem{Berntorp2014models}
K.~Berntorp, B.~Olofsson, K.~Lundahl, and L.~Nielsen, ``Models and methodology
  for optimal trajectory generation in safety-critical road--vehicle
  manoeuvres,'' {\em Vehicle System Dynamics}, vol.~52, no.~10, pp.~1304--1332,
  2014.

\bibitem{Li2015unified}
B.~Li and Z.~Shao, ``A unified motion planning method for parking an autonomous
  vehicle in the presence of irregularly placed obstacles,'' {\em
  Knowledge-Based Systems}, vol.~86, pp.~11--20, 2015.

\bibitem{Li2016time}
B.~Li, K.~Wang, and Z.~Shao, ``Time-optimal maneuver planning in automatic
  parallel parking using a simultaneous dynamic optimization approach,'' {\em
  IEEE Transactions on Intelligent Transportation Systems}, vol.~17, no.~11,
  pp.~3263--3274, 2016.

\bibitem{Zhang2018autonomous}
X.~Zhang, A.~Liniger, A.~Sakai, and F.~Borrelli, ``Autonomous parking using
  optimization-based collision avoidance,'' in {\em 2018 IEEE Conference on
  Decision and Control (CDC)}, pp.~4327--4332, IEEE, 2018.

\bibitem{Sun2021successive}
C.~Sun, Q.~Li, B.~Li, and L.~Li, ``A successive linearization in feasible set
  algorithm for vehicle motion planning in unstructured and low-speed
  scenarios,'' {\em IEEE Transactions on Intelligent Transportation Systems},
  vol.~23, no.~4, pp.~3724--3736, 2021.

\bibitem{Li2021optimization}
B.~Li, T.~Acarman, Y.~Zhang, Y.~Ouyang, C.~Yaman, Q.~Kong, X.~Zhong, and
  X.~Peng, ``Optimization-based trajectory planning for autonomous parking with
  irregularly placed obstacles: A lightweight iterative framework,'' {\em IEEE
  Transactions on Intelligent Transportation Systems}, vol.~23, no.~8,
  pp.~11970--11981, 2021.

\bibitem{Han2023efficient}
Z.~Han, Y.~Wu, T.~Li, L.~Zhang, L.~Pei, L.~Xu, C.~Li, C.~Ma, C.~Xu, S.~Shen,
  {\em et~al.}, ``An efficient spatial-temporal trajectory planner for
  autonomous vehicles in unstructured environments,'' {\em IEEE Transactions on
  Intelligent Transportation Systems}, 2023.

\bibitem{Zhu2015convex}
Z.~Zhu, E.~Schmerling, and M.~Pavone, ``A convex optimization approach to
  smooth trajectories for motion planning with car-like robots,'' in {\em 2015
  54th IEEE conference on decision and control (CDC)}, pp.~835--842, IEEE,
  2015.

\bibitem{Liu2017convex}
C.~Liu, C.-Y. Lin, Y.~Wang, and M.~Tomizuka, ``Convex feasible set algorithm
  for constrained trajectory smoothing,'' in {\em 2017 American Control
  Conference (ACC)}, pp.~4177--4182, IEEE, 2017.

\bibitem{Liu2018convex}
C.~Liu, C.-Y. Lin, and M.~Tomizuka, ``The convex feasible set algorithm for
  real time optimization in motion planning,'' {\em SIAM Journal on Control and
  optimization}, vol.~56, no.~4, pp.~2712--2733, 2018.

\bibitem{Deits2015computing}
R.~Deits and R.~Tedrake, ``Computing large convex regions of obstacle-free
  space through semidefinite programming,'' in {\em Algorithmic Foundations of
  Robotics XI: Selected Contributions of the Eleventh International Workshop on
  the Algorithmic Foundations of Robotics}, pp.~109--124, Springer, 2015.

\bibitem{Liu2017planning}
S.~Liu, M.~Watterson, K.~Mohta, K.~Sun, S.~Bhattacharya, C.~J. Taylor, and
  V.~Kumar, ``Planning dynamically feasible trajectories for quadrotors using
  safe flight corridors in 3-d complex environments,'' {\em IEEE Robotics and
  Automation Letters}, vol.~2, no.~3, pp.~1688--1695, 2017.

\bibitem{Bergman2018combining}
K.~Bergman and D.~Axehill, ``Combining homotopy methods and numerical optimal
  control to solve motion planning problems,'' in {\em 2018 IEEE Intelligent
  Vehicles Symposium (IV)}, pp.~347--354, IEEE, 2018.

\bibitem{Shi2019bilevel}
S.~Shi, Y.~Xiong, J.~Chen, and C.~Xiong, ``A bilevel optimal motion planning
  (bomp) model with application to autonomous parking,'' {\em International
  Journal of Intelligent Robotics and Applications}, vol.~3, no.~4,
  pp.~370--382, 2019.

\bibitem{Zhou2020autonomous}
J.~Zhou, R.~He, Y.~Wang, S.~Jiang, Z.~Zhu, J.~Hu, J.~Miao, and Q.~Luo,
  ``Autonomous driving trajectory optimization with dual-loop iterative
  anchoring path smoothing and piecewise-jerk speed optimization,'' {\em IEEE
  Robotics and Automation Letters}, vol.~6, no.~2, pp.~439--446, 2020.

\bibitem{Gilbert2002fast}
E.~G. Gilbert, D.~W. Johnson, and S.~S. Keerthi, ``A fast procedure for
  computing the distance between complex objects in three-dimensional space,''
  {\em IEEE Journal on Robotics and Automation}, vol.~4, no.~2, pp.~193--203,
  2002.

\bibitem{Frison2020hpipm}
G.~Frison and M.~Diehl, ``Hpipm: a high-performance quadratic programming
  framework for model predictive control,'' {\em IFAC-PapersOnLine}, vol.~53,
  no.~2, pp.~6563--6569, 2020.

\end{thebibliography}
\end{document}